\documentclass[runningheads]{llncs}

\usepackage{eccv}
\usepackage{svg}
\usepackage{graphicx}
\usepackage{booktabs}
\usepackage{amsmath,amssymb}
\usepackage{multirow}
\usepackage{xcolor}
\usepackage{pifont}
\usepackage[accsupp]{axessibility}
\usepackage[breaklinks,colorlinks,citecolor=blue]{hyperref}
\usepackage{float}
\usepackage[section]{placeins}
\usepackage{orcidlink}

\newcommand{\method}{AlignJEPA}

\begin{document}

% ---------------------------------------------------------------
\title{AlignJEPA: Predictive Vision-Language Alignment for Remote Sensing Foundation Models}

\titlerunning{AlignJEPA: Predictive Vision-Language Alignment}

\author{
Md Aminur Hossain\inst{1}\orcidlink{0009-0003-6357-7480} \and
Omkumar Vaghasiya\inst{3}\orcidlink{0009-0005-8194-2503} \and
Rajeev Ranjan Dwivedi\inst{3}\orcidlink{0000-0002-2127-564X} \and
Vinod Kurmi\inst{3}\orcidlink{0000-0003-0332-5647} \and
Biplab Banerjee\inst{2}\orcidlink{0000-0001-8371-8138}
}

\authorrunning{M.A. Hossain et al.}

\institute{
Space Applications Centre, ISRO, Ahmedabad, India
\and
CSRE, Indian Institute of Technology Bombay, India\\
\and
Indian Institute of Science Education and Research (IISER) Bhopal, India\\
\email{\{md.aminurhossain, vaghasiyaom6, getbiplab\}@gmail.com}}

\maketitle

\begin{abstract}
Remote sensing (RS) foundation models provide transferable Earth observation representations across sensors, resolutions, and geographies, yet most remain weakly aligned with natural language, limiting natural-language archive search, image--text retrieval, and question-conditioned analysis. We propose \method, a JEPA-inspired predictive vision-language alignment framework for remote sensing foundation models. \method{} uses a pretrained AnySat visual encoder and a RemoteCLIP text encoder while training only a lightweight predictive alignment network. Instead of relying on global image--text contrastive alignment alone, the framework predicts remote-sensing text embeddings from masked visual foundation-model tokens. Its mask-aware multi-scale predictive aligner aggregates visible tokens at fine, regional, and global scales, jointly models them with a cross-scale Transformer, and projects the resulting representation into the text space using learned query pooling. Training combines semantic prediction with bidirectional contrastive retrieval. We train and evaluate \method{} on BigEarthNet.txt for natural-language Sentinel retrieval, evaluate cross-dataset adaptation on RSICD, and use RSVQA only as a closed-set representation probe. \method{} provides a parameter-efficient route for aligning Earth observation foundation models with language.

\keywords{RS foundation models \and Vision-language alignment \and Joint-Embedding Predictive Architecture \and Retrieval \and Geospatial AI}
\end{abstract}

% ---------------------------------------------------------------
\section{Introduction}

Earth observation (EO) foundation models have become central to scalable remote-sensing representation learning. Masked image modelling methods such as SatMAE and Scale-MAE show that self-supervised pretraining can improve transfer across EO tasks while reducing dependence on task-specific labels~\cite{satmae,scalemae}. Beyond optical imagery, CROMA learns radar--optical representations from Sentinel-1/Sentinel-2 observations~\cite{croma}, while DOFA studies sensor-adaptive modelling across heterogeneous spectral configurations~\cite{dofa}. More recent EO encoders, including AnySat and Prithvi-EO-2.0, further demonstrate the value of large-scale pretraining for land-cover mapping, segmentation, and multisensor classification~\cite{anysat,prithvi2}. However, many EO foundation models remain primarily visual and are not directly aligned with natural language.

Natural-language interaction is increasingly important for practical geospatial analysis. Users may want to retrieve satellite imagery using descriptions such as ``cropland with forest patches and nearby water bodies'', compare regions through textual concepts, or ask questions about land-cover composition and scene context. Remote-sensing vision-language models address this need by aligning satellite images with text. RemoteCLIP adapts CLIP-style contrastive learning to remote-sensing image--text pairs~\cite{remoteclip}, and GeoRSCLIP scales geospatial image--text pretraining for retrieval and zero-shot recognition~\cite{georsclip}. Instruction-oriented models such as GeoChat and SkyEyeGPT further extend RS vision-language learning toward dialogue and visual question answering~\cite{geochat,skyeyegpt}. Most existing RS-VLM encoders, however, rely mainly on global image--text alignment and do not explicitly exploit the spatial token representations produced by pretrained EO foundation models such as AnySat.

This paper proposes \method, a JEPA-inspired predictive vision-language alignment framework for connecting EO visual representations with a remote-sensing language embedding space. Given an EO image, a pretrained AnySat encoder extracts spatial visual tokens, while a RemoteCLIP text encoder defines the target language-semantic space. A lightweight predictive alignment network receives partially visible visual tokens and predicts the corresponding text-space representation. To better exploit spatial context under masking, we introduce a mask-aware multi-scale predictive aligner that aggregates visible tokens at fine, regional, and global scales before cross-scale Transformer prediction. Following the predictive spirit of JEPA, the model learns from semantic latent targets rather than pixel reconstruction~\cite{ijepa,vjepa}. Unlike image-only JEPA models, the prediction target in \method{} is a frozen remote-sensing text embedding, enabling language-aligned EO representations for retrieval and probing tasks.

We use BigEarthNet.txt as the main benchmark because it provides large-scale image--text supervision over co-registered Sentinel-1/Sentinel-2 imagery~\cite{bigearthnettxt}. We train \method{} only with image--caption pairs, leaving its VQA and referring-expression annotations unused to keep the objective focused on natural-language Sentinel retrieval. We further evaluate cross-dataset adaptation on RSICD~\cite{rsicd}
and closed-set representation probing on RSVQA~\cite{rsvqa}. In all experiments, the visual and text encoders remain frozen; only the alignment network or task-specific probe head is trained.

The main contributions are:
\begin{itemize}
    \item We propose \method{}, a JEPA-inspired framework for predictively aligning pretrained EO visual tokens with a remote-sensing language embedding space.
    \item We introduce a mask-aware multi-scale predictive aligner that aggregates visible EO tokens across fine, regional, and global spatial scales and jointly models their interactions for text-space semantic prediction.
    \item We keep both EO visual and text encoders frozen, making the approach parameter-efficient.
    \item We evaluate \method{} on BigEarthNet.txt retrieval, BigEarthNet.txt-to-RSICD adaptation, and closed-set RSVQA probing.
\end{itemize}

% ---------------------------------------------------------------

\section{Related Work}

\subsection{Remote Sensing Foundation Models}

Remote-sensing foundation models learn transferable EO representations from large satellite archives. SatMAE adapts masked autoencoding to multispectral imagery, while Scale-MAE studies scale-aware pretraining for geospatial scenes~\cite{satmae,scalemae}. Multisensor models extend this direction: CROMA combines radar--optical masked modelling with cross-modal contrastive learning~\cite{croma}, and DOFA introduces wavelength-conditioned adaptation for different sensor configurations~\cite{dofa}. AnySat is especially relevant to our work because it uses resolution-adaptive spatial encoders and a JEPA-based objective to produce rich EO visual tokens across modalities and resolutions~\cite{anysat}.

Recent large-scale models further emphasize temporal, multisensor, and globally distributed pretraining. SkySense and Prithvi-EO-2.0 explore spatiotemporal and multi-temporal EO representation learning~\cite{skysense,prithvi2}, while TerraFM and MMEarth focus on scalable Sentinel-based pretraining and multimodal EO resources~\cite{terrafm,mmearth}. Building on such pretrained EO representations, \method{} learns a predictive alignment network that exploits visual tokens at multiple scales and maps them into a remote-sensing language embedding space.

\subsection{Remote Sensing Vision-Language Models}

Vision-language learning is increasingly important for remote sensing because many EO queries are naturally expressed in text. CLIP introduced large-scale contrastive image--text pretraining in the natural-image domain~\cite{clip}, but direct transfer to satellite imagery is limited by domain shift and remote-sensing-specific semantics. RemoteCLIP addresses this gap through RS-specific image--text supervision for zero-shot classification and retrieval~\cite{remoteclip}. GeoRSCLIP further scales remote-sensing vision-language pretraining using RS5M and studies retrieval and zero-shot recognition in the geospatial domain~\cite{georsclip}.

Other works explore alternative supervision and interaction settings. Ground-Remote Alignment connects satellite images with language through co-located ground-level imagery~\cite{groundremote}, FLAVARS combines contrastive and masked-modelling objectives~\cite{flavars}, and LRSCLIP studies alignment with longer remote-sensing text~\cite{lrsclip}. Instruction-tuned models such as GeoChat, SkyEyeGPT, EarthGPT, and SkySenseGPT target dialogue, VQA, grounding, and multi-task reasoning~\cite{geochat,skyeyegpt,earthgpt,skysensegpt}. These methods improve RS vision-language understanding, but most either rely on global image--text alignment or require generative instruction tuning. \method{} instead focuses on retrieval-oriented representation learning by predicting frozen text-space embeddings from masked, multi-scale EO visual context.

\subsection{Predictive Joint-Embedding Learning}

Joint-Embedding Predictive Architectures learn representations by predicting latent targets from context rather than reconstructing low-level observations~\cite{hjepa}. I-JEPA applies this principle to images by predicting target-block representations from visible context and shows that latent prediction can encourage semantic representation learning~\cite{ijepa}. V-JEPA extends the same idea to video through masked spatiotemporal prediction~\cite{vjepa}. \method{} adopts this predictive philosophy but changes the target space: instead of predicting missing visual latents, it predicts a frozen remote-sensing text embedding from partially visible EO visual tokens. This makes \method{} a JEPA-inspired vision-language alignment framework rather than a conventional image-only JEPA model. In contrast to CLIP-style global alignment and instruction-tuned RS-VLMs, \method{} uses frozen EO and language encoders and learns only a mask-aware multi-scale predictive adapter that maps spatial EO context into the remote-sensing text-embedding space.
\section{Proposed AlignJEPA Framework}

\subsection{Overview}

\begin{figure}[t]
\centering
\includegraphics[width=1\linewidth]{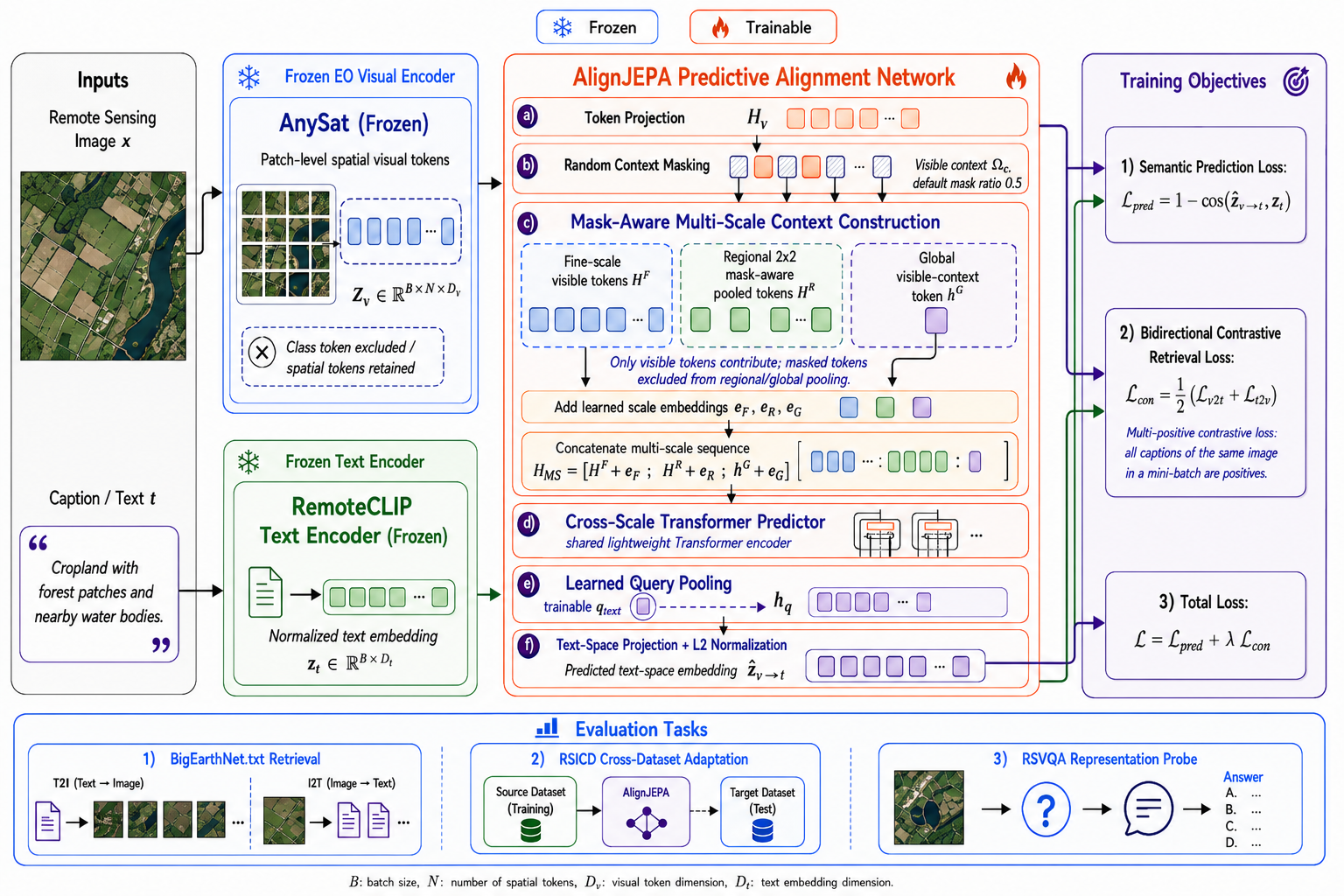}
\caption{
Overview of \method. A pretrained EO visual encoder extracts spatial tokens from remote-sensing imagery, and a RemoteCLIP text encoder provides the target language-semantic embedding. The trainable predictive alignment network maps partially visible visual tokens into the text embedding space using semantic prediction and bidirectional contrastive retrieval objectives.
}
\label{fig:overview}
\end{figure}

The proposed \method{} framework aligns pretrained visual representations of EO with a remote-sensing language embedding space through a trainable predictive alignment network. Given an image--text pair $(x,t)$, the EO visual encoder extracts spatial visual tokens from the remote-sensing image, while the text encoder maps the associated text into a language-semantic embedding. The visual and text encoder parameters remain unchanged during training; only the predictive alignment network is optimized. The network receives partially visible visual tokens and predicts the corresponding text-space representation, encouraging language-level semantic alignment from incomplete geospatial evidence. Rather than processing visible evidence only at its native token scale, the predictive aligner constructs fine-, regional-, and global-context representations using only visible tokens. A shared Transformer then models interactions across these scales before predicting the target language embedding. Fig.~\ref{fig:overview} illustrates the architecture.

Although \method{} is inspired by JEPA, it differs from image-only JEPA models in its target definition. Conventional JEPA methods predict missing visual latents from visible context~\cite{ijepa,vjepa}. In contrast, \method{} predicts a fixed remote-sensing text embedding from masked EO visual tokens, transferring the predictive principle of JEPA to vision-language alignment.

\subsection{Problem Formulation}
\label{sec:problem-formulation}

Let $x$ denote a remote-sensing image and $t$ denote its associated text. The EO visual encoder $E_v$ produces a sequence of visual tokens:
\begin{equation}
Z_v = E_v(x), \quad Z_v \in \mathbb{R}^{B \times N \times D_v},
\end{equation}
where $B$ is the batch size, $N$ is the number of visual tokens, and $D_v$ is the visual token dimension.

The text encoder $E_t$ maps $t$ to a language-semantic embedding:
\begin{equation}
z_t = E_t(t), \quad z_t \in \mathbb{R}^{B \times D_t},
\end{equation}
which is L2-normalized:
\begin{equation}
z_t = \frac{z_t}{\|z_t\|_2}.
\end{equation}

The goal of \method{} is to learn a predictor $P_{\theta}$ that maps partially visible visual evidence into the text embedding space:
\begin{equation}
\hat{z}_{v \rightarrow t}
=
P_{\theta}(Z_v,\Omega_c),
\end{equation}
where $\Omega_c$ denotes the visible context-token indices after masking and retains the spatial visibility information required for multi-scale context construction. The notation $\hat{z}_{v \rightarrow t}$ indicates that the prediction is produced from the visual branch but lies in the text embedding space. During training, $\Omega_c$ is sampled according to the masking policy. During retrieval and representation probing, masking is disabled and all visual tokens are used to compute the image-side representation, while $z_t$ is used as the text-side representation.

\subsection{EO Visual Token Encoder}

We use AnySat as the EO visual encoder because it supports heterogeneous remote-sensing inputs and produces spatial token representations suitable for predictive learning~\cite{anysat}. During \method{} training, the visual encoder parameters remain unchanged. In the main BigEarthNet.txt setting, Sentinel-2 imagery is encoded into spatial visual tokens and aligned with caption embeddings.

\subsection{Remote-Sensing Text Encoder}

We use the RemoteCLIP text encoder as the language branch~\cite{remoteclip}. RemoteCLIP provides a remote-sensing-aware text embedding space, making it suitable as a semantic target for image--text alignment. For a text input $t$, the encoder produces the normalized embedding $z_t$ defined in Sec.~\ref{sec:problem-formulation}. The text encoder parameters remain unchanged throughout training, keeping the language-semantic target space stable.

\begin{figure}[h]
\vspace{-2em}
    \centering
    \includegraphics[width=1.0\linewidth]{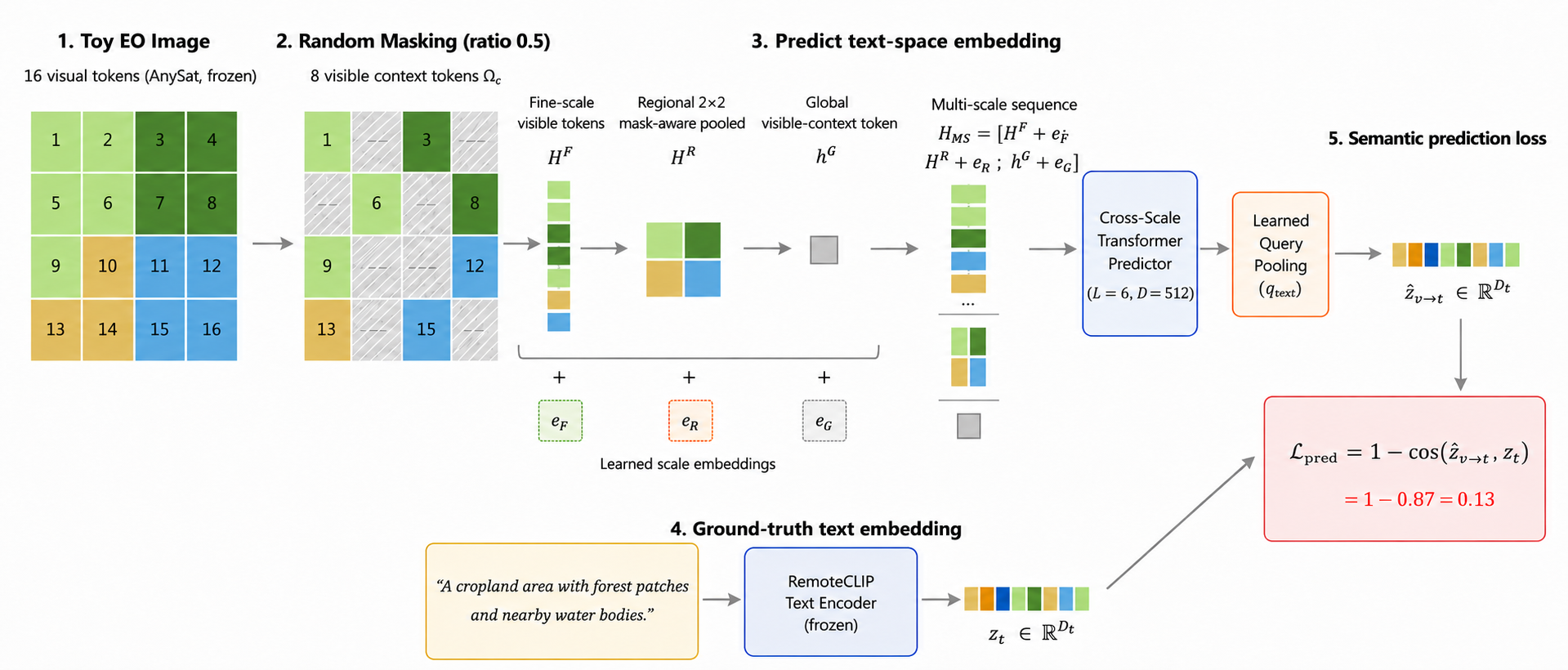}
   \caption{Toy $4{\times}4$ example of AlignJEPA: partially visible EO context
is mapped into the RemoteCLIP text-embedding space and aligned with
the caption embedding $z_t$ via $\mathcal{L}_{\text{pred}}$.}
    \label{fig:toy_example} \vspace{-1em}
\end{figure}

Fig.~\ref{fig:toy_example} illustrates one forward pass of AlignJEPA on a toy $4{\times}4$ scene with cropland, forest, soil, and water. Half of the 16 visual tokens are masked, and the visible context $\Omega_c$ is used to construct fine-, regional-, and global-context representations. These multi-scale representations are processed by the cross-scale Transformer predictor and learned query pooling module to produce $\hat{z}_{v\to t}$ in the RemoteCLIP text space. In parallel, the frozen text encoder maps the caption to $z_t$, and
$\mathcal{L}_{\text{pred}} = 1 - \cos(\hat{z}_{v\to t}, z_t)$ pulls the
two embeddings together. The example makes explicit that prediction happens \emph{in the text space}, not in pixel or visual-latent space.

\subsection{\method{} Predictive Alignment Network}

The predictive alignment network is the trainable component of \method{}. It transforms EO visual tokens into a RemoteCLIP-compatible text-space embedding through token projection, visual context masking, mask-aware multi-scale context construction, cross-scale Transformer prediction, learned query pooling, and text-space projection.

\subsubsection{Token Projection}

The EO visual tokens are first projected into a common hidden dimension $D$:
\begin{equation}
H_v = \phi_{\mathrm{proj}}(Z_v), \quad H_v \in \mathbb{R}^{B \times N \times D},
\end{equation}
where $\phi_{\mathrm{proj}}$ is implemented using a linear layer followed by LayerNorm and a non-linear activation.

\subsubsection{Visual Context Masking}

To introduce predictive learning, we mask a subset of visual tokens before prediction. Let $\Omega_c$ denote the set of visible context-token indices, and let $m_i=\mathbf{1}[i\in\Omega_c]$ denote the corresponding binary visibility indicator. The fine-scale visible tokens are $H^{F}=H_v[\Omega_c]$, while their original spatial indices are retained for multi-scale context construction. The remaining tokens are not reconstructed directly. Instead, the network predicts a language-semantic target from the visible context, following the JEPA-style principle of inferring latent targets from context rather than reconstructing pixels~\cite{ijepa,vjepa}. We use a default mask ratio of $0.5$ and study masking variants in the ablation analysis.

\subsubsection{Mask-Aware Multi-Scale Context}

To preserve both local evidence and broader scene context under masking, we construct fine, regional, and global representations using only visible tokens. The fine-scale representation is the set of visible projected tokens $H^{F}$. For each non-overlapping $2{\times}2$ spatial region $r$ containing at least one visible token, a regional token is obtained by mask-aware pooling:
\begin{equation}
h_{r}^{R}
=
\frac{
\sum_{i\in r} m_i h_i
}{
\sum_{i\in r} m_i+\epsilon
},
\end{equation}
where $h_i$ denotes the projected token at spatial location $i$. Collecting the valid regional tokens gives
$H^{R}=\{h_r^{R}\}_{r\in\mathcal{R}_v}$,
where $\mathcal{R}_v$ denotes the set of regional windows containing at least one visible token.

A global visible-context token is computed as
\begin{equation}
h^{G}
=
\frac{
\sum_{i=1}^{N} m_i h_i
}{
\sum_{i=1}^{N} m_i+\epsilon
}.
\end{equation}
Masked tokens therefore do not contribute to either regional or global pooling, preventing information leakage through coarse-scale representations. Learned scale embeddings $e_F$, $e_R$, and $e_G$ identify the three context levels, and the multi-scale sequence is formed as
\begin{equation}
H_{\mathrm{MS}}
=
\left[
H^{F}+e_F;\;
H^{R}+e_R;\;
h^{G}+e_G
\right].
\end{equation}

\subsubsection{Cross-Scale Transformer Prediction}

The multi-scale context sequence is processed by a shared lightweight Transformer encoder~\cite{transformer}:
\begin{equation}
H_p = \phi_{\mathrm{trans}}(H_{\mathrm{MS}}),
\end{equation}
where self-attention jointly models interactions among fine local evidence, regional structure, and scene-level context.

\subsubsection{Learned Query Pooling}

To obtain a single representation for text-space prediction, we use a learnable query token, similar in spirit to query-based visual pooling in vision-language models~\cite{blip2}:
\begin{equation}
h_q = \mathrm{Attn}(q_{\mathrm{text}},H_p,H_p),
\end{equation}
where $q_{\mathrm{text}}$ is a trainable query vector independent of the input caption and $h_q$ is the pooled semantic representation.

\subsubsection{Text-Space Projection}

The pooled representation is projected into the RemoteCLIP text embedding dimension:
$\hat{z}_{v \rightarrow t} = \phi_{\mathrm{text}}(h_q)$,
where $\phi_{\mathrm{text}}$ is a projection head. The predicted embedding is L2-normalized:
\begin{equation}
\hat{z}_{v \rightarrow t}
=
\frac{\hat{z}_{v \rightarrow t}}
{\|\hat{z}_{v \rightarrow t}\|_2}.
\end{equation}

\subsection{Training Objective}

The training objective combines semantic prediction with contrastive retrieval alignment, following CLIP-style image--text retrieval formulations~\cite{clip}. The prediction loss aligns the visual-to-text embedding with the target text embedding, while the contrastive loss improves retrieval discrimination.

\subsubsection{Semantic Prediction Loss}

The semantic prediction loss is computed over a mini-batch:
\begin{equation}
\mathcal{L}_{\mathrm{pred}}
=
\frac{1}{B}
\sum_{i=1}^{B}
\left(
1 -
\hat{z}_{v \rightarrow t,i}^{\top} z_{t,i}
\right),
\end{equation}
where $\hat{z}_{v \rightarrow t,i}$ is the predicted text-space embedding from the visual branch and $z_{t,i}$ is the corresponding RemoteCLIP text embedding. Since both embeddings are L2-normalized, the dot product is equivalent to cosine similarity.

\subsubsection{Contrastive Retrieval Loss}

For a batch of image--text pairs, we compute:
\begin{equation}
S_{ij}
=
\frac{
\hat{z}_{v \rightarrow t,i}^{\top} z_{t,j}
}
{\tau},
\end{equation}
where $\tau$ is a temperature parameter.

Since each image may have multiple valid captions or related text descriptions, we use a multi-positive contrastive formulation. In BigEarthNet.txt, all captions associated with the same image identifier are treated as positives. Thus, $\mathcal{P}(i)$ contains all text entries in the mini-batch corresponding to image $i$, and $\mathcal{Q}(j)$ contains all image entries associated with the same underlying image as text $j$. When only one caption for an image appears in a mini-batch, this reduces to standard bidirectional InfoNCE. Let $\mathcal{P}(i)$ denote the set of positive text indices associated with image $i$. The image-to-text loss is:
\begin{equation}
\mathcal{L}_{v2t}
=
-\frac{1}{B}
\sum_{i=1}^{B}
\log
\frac{
\sum_{j \in \mathcal{P}(i)} \exp(S_{ij})
}
{
\sum_{j=1}^{B} \exp(S_{ij})
}.
\end{equation}
Similarly, let $\mathcal{Q}(j)$ denote the set of positive image indices associated with text $j$. The text-to-image loss is:
\begin{equation}
\mathcal{L}_{t2v}
=
-\frac{1}{B}
\sum_{j=1}^{B}
\log
\frac{
\sum_{i \in \mathcal{Q}(j)} \exp(S_{ij})
}
{
\sum_{i=1}^{B} \exp(S_{ij})
}.
\end{equation}
The bidirectional retrieval loss is:
\begin{equation}
\mathcal{L}_{\mathrm{con}}
=
\frac{1}{2}
\left(
\mathcal{L}_{v2t}
+
\mathcal{L}_{t2v}
\right).
\end{equation}

\subsubsection{Total Objective}

The final training objective is:
\begin{equation}
\mathcal{L}
=
\mathcal{L}_{\mathrm{pred}}
+
\lambda
\mathcal{L}_{\mathrm{con}},
\end{equation}
where $\lambda$ controls the contribution of the contrastive retrieval loss. Unless otherwise stated, we set $\lambda=1.0$ and $\tau=0.07$.
% ---------------------------------------------------------------
\section{Experiments}

\subsection{Evaluation Setting}

We evaluate \method{} as a parameter-efficient alignment framework for mapping pretrained EO visual representations to remote-sensing language embeddings. BigEarthNet.txt is the main training and evaluation benchmark, using
image--caption pairs from its co-registered Sentinel-1/Sentinel-2 archive,
with Sentinel-2 used as the visual input~\cite{bigearthnettxt}. We also evaluate BigEarthNet.txt-to-RSICD adaptation for RGB aerial
retrieval~\cite{rsicd} and use RSVQA as a closed-set representation probe for question-conditioned prediction~\cite{rsvqa}. For RSICD, the predictive alignment network pretrained on BigEarthNet.txt is fine-tuned on the RSICD training split, while the EO visual and text encoders remain frozen. In all experiments, only the alignment network or task-specific probe head is optimized.

\subsection{Datasets}

\paragraph{BigEarthNet.txt} is the primary dataset in this work. It extends the BigEarthNet Sentinel archive with natural-language supervision and contains co-registered Sentinel-1 and Sentinel-2 imagery paired with text annotations~\cite{bigearthnettxt}. The full dataset includes caption-style descriptions, question--answer pairs, and referring-expression-style instructions. In this paper, we use only image--caption pairs as training supervision and evaluate natural-language Sentinel image retrieval. We do not use VQA pairs or referring-expression instructions for training \method{}. To avoid leakage across multiple textual descriptions of the same image, all annotations belonging to the same image are assigned to the same split. Unless otherwise stated, we use an image-level split of approximately $80\%$ for training, $10\%$ for validation, and $10\%$ for testing.

\vspace{0.5em}\noindent RSICD is used for cross-dataset image--text retrieval adaptation~\cite{rsicd}. It contains RGB aerial images paired with natural-language captions and differs from BigEarthNet.txt in sensor characteristics, spatial resolution, image style, and annotation distribution. The predictive alignment network initialized from BigEarthNet.txt training is
fine-tuned on the RSICD training split and evaluated on the corresponding test split using the same retrieval metrics. RSICD RGB images are processed using the RGB-compatible pathway of the AnySat encoder.

\vspace{0.5em}\noindent RSVQA is used as a representation probing benchmark for question-conditioned prediction~\cite{rsvqa}. We do not treat this experiment as generative visual question answering. Instead, we combine the learned image-side representation with a frozen question embedding and train a lightweight answer classifier. This setting measures whether the aligned representation contains information useful for language-conditioned downstream prediction. In our experiments, we evaluate on both RSVQA-LR and RSVQA-HR, reporting overall accuracy together with answer-type accuracy.

\subsection{Tasks and Metrics}

For image--text retrieval, we evaluate both text-to-image and image-to-text
directions. We report Recall@1, Recall@5, and Recall@10, which are standard
metrics for cross-modal retrieval in remote sensing and vision-language models.
During retrieval evaluation, masking is disabled and all test samples are used
to form the retrieval galleries. Captions associated with the same image are
treated as valid matches when computing Recall@K.
For RSVQA, we report overall answer accuracy and, where applicable, accuracy
by answer type. Across all tables, bold text indicates the best performance
within the corresponding comparison block.

\subsection{Baselines}

We compare \method{} with released RS vision--language models, AnySat-based
alignment baselines, and ablations. RemoteCLIP and GeoRSCLIP provide external
image--text retrieval references using their released encoders~\cite{remoteclip,georsclip}.
To isolate predictive visual-to-text alignment, we include two AnySat baselines.
\textbf{AnySat + linear projection} maps pooled AnySat tokens to the text space
with a trained linear layer, while \textbf{AnySat + contrastive alignment} optimizes a trainable aligner using only the bidirectional contrastive retrieval loss.

We also report two direct ablations: \textbf{\method{} without masking}, which
removes visual context masking, and \textbf{\method{} without semantic
prediction}, which removes $\mathcal{L}_{\mathrm{pred}}$ and retains only
contrastive supervision. The full \method{} combines the mask-aware multi-scale
predictive aligner with semantic prediction and bidirectional contrastive retrieval.

For RSVQA, we compare representation probes based on RemoteCLIP, AnySat,
contrastive alignment, and \method{}. In all variants, only the lightweight
answer head is trained on the RSVQA training split, while the visual and text
encoders remain frozen.

\subsection{Implementation Details}

We use the publicly released pretrained AnySat base checkpoint as the EO visual
encoder and RemoteCLIP as the text encoder. AnySat supports heterogeneous
resolutions, scales, and modalities, including aerial imagery, Sentinel-1/2,
and Landsat~\cite{anysat}, making it suitable for our Sentinel-2
BigEarthNet.txt and RGB RSICD settings. We extract patch-level spatial
representations, exclude the class token, and retain the native 2D patch-grid
organization for $2\times2$ regional pooling. Both encoders remain frozen;
only the predictive alignment network is trained. At inference, masking is
disabled and all visual tokens are used.

Training uses AdamW with learning rate $1\times10^{-4}$, weight decay $0.05$,
cosine decay, $5\%$ warm-up, gradient clipping at $1.0$, and mixed precision.
We use a mask ratio of $0.5$, contrastive temperature $\tau=0.07$, and loss
weight $\lambda=1.0$.

The predictive aligner comprises token projection, mask-aware multi-scale
context construction, a shared cross-scale Transformer, learned query pooling,
and text-space projection. Fine-scale visible tokens, non-overlapping
$2\times2$ regional pooling, and one global context token form the three scales,
distinguished by learned scale embeddings. We use $D=512$, $L=6$ Transformer
layers, $H=8$ attention heads, and dropout $0.1$. Learned query pooling uses one
trainable query independent of the caption, followed by projection to the
RemoteCLIP text dimension. Training runs for $200$ epochs with global batch
size $256$ and $224\times224$ inputs on $4$ NVIDIA A100 GPUs. Models are
selected by validation mean recall on BigEarthNet.txt, and the predictive
aligner has approximately $20$M trainable parameters. For RSICD adaptation, we
initialize from the BigEarthNet.txt checkpoint, fine-tune with the same
objective, and select models by RSICD validation mean recall.

\section{Results and Analysis}

\subsection{BigEarthNet.txt Retrieval}

Table~\ref{tab:bige-retrieval} reports image--text retrieval results on BigEarthNet.txt using image--caption supervision. RemoteCLIP and GeoRSCLIP are used as released RS vision-language references, while the AnySat-based baselines isolate the effect of aligning frozen EO visual tokens with the RemoteCLIP text space.

\begin{table*}[ht]
\centering
\caption{Image--text retrieval results on BigEarthNet.txt. T2I denotes text-to-image retrieval and I2T denotes image-to-text retrieval.}
\label{tab:bige-retrieval}
\resizebox{\linewidth}{!}{
\begin{tabular}{lcccccc}
\toprule
Method & T2I R@1 & T2I R@5 & T2I R@10 & I2T R@1 & I2T R@5 & I2T R@10 \\
\midrule
RemoteCLIP zero-shot & 12.42 & 33.80 & 46.90 & 13.68 & 35.16 & 48.60 \\
GeoRSCLIP zero-shot & 13.61 & 35.42 & 49.20 & 14.52 & 37.23 & 50.40 \\
AnySat + Linear Projection & 10.79 & 30.52 & 43.74 & 11.67 & 31.83 & 44.59 \\
AnySat + Contrastive Alignment & 16.90 & 40.60 & 55.35 & 17.80 & 42.42 & 57.13 \\
\textbf{\method{}} & \textbf{20.72} & \textbf{47.81} & \textbf{62.09} & \textbf{21.89} & \textbf{49.31} & \textbf{63.64} \\
\bottomrule
\end{tabular}
}
\end{table*}

The full \method{} model achieves the best performance across all BigEarthNet.txt retrieval metrics. Compared with AnySat + Contrastive Alignment, \method{} improves both T2I and I2T retrieval, suggesting that masked multi-scale visual-to-text prediction provides complementary supervision beyond contrastive learning alone. The component-level effects of masking and semantic prediction are analyzed separately in Table~\ref{tab:ablation-summary}.

\begin{table*}[h]
\centering
\caption{Image--text retrieval on RSICD. Published RSICD results are shown
only for context because their training protocols differ. The bottom block
reports our controlled cross-dataset adaptation setting, where the
BigEarthNet.txt-trained alignment network is fine-tuned on the RSICD training
split while AnySat and RemoteCLIP remain frozen.}
\label{tab:rsicd}
\footnotesize
\scalebox{0.85}{
\begin{tabular}{lcccccc}
\toprule
\multirow{2}{*}{Method} & \multicolumn{3}{c}{Text-to-Image (T2I)} & \multicolumn{3}{c}{Image-to-Text (I2T)} \\
\cmidrule(lr){2-4} \cmidrule(lr){5-7}
 & R@1 & R@5 & R@10 & R@1 & R@5 & R@10 \\
\midrule
VSE++~\cite{vsepp} & 2.82 & 11.32 & 18.10 & 3.38 & 9.51 & 17.46 \\
SCAN~\cite{scan} & 3.91 & 16.20 & 26.49 & 4.39 & 10.90 & 17.64 \\
CAMP~\cite{camp} & 4.15 & 15.23 & 27.81 & 5.12 & 12.89 & 21.12 \\
MTFN~\cite{mtfn} & 4.90 & 17.17 & 29.49 & 5.02 & 12.52 & 19.74 \\
KCR~\cite{kcr} & 4.76 & 18.59 & 27.20 & 5.84 & 22.31 & 36.12 \\
SWAN~\cite{swan} & 5.56 & 22.26 & 37.41 & 7.41 & 20.13 & 30.86 \\
HVSA~\cite{hvsa} & 5.51 & 21.13 & 34.13 & 7.47 & 20.62 & 32.11 \\
FAAMI~\cite{faami} & 8.11 & 25.59 & 41.37 & 10.44 & 22.66 & 30.89 \\
Multilanguage~\cite{multilanguage} & 9.14 & 28.96 & 44.59 & 10.70 & 29.64 & 41.53 \\
PE-RSITR~\cite{persitr} & 11.63 & 33.92 & 50.73 & 14.13 & 31.51 & 44.78 \\
MTGFE~\cite{mtgfe} & 8.67 & 27.56 & 43.92 & 15.28 & 37.05 & 51.60 \\
CLIP-RSICD~\cite{cliprsicd} & 11.16 & 33.25 & 48.91 & 14.09 & 30.10 & 43.64 \\
CLIP-Cap-4~\cite{rscapret} & 13.83 & 39.07 & 56.05 & 17.02 & 33.94 & 47.76 \\
RemoteCLIP~\cite{remoteclip} & 14.73 & 39.93 & 56.58 & 18.39 & 37.42 & 51.05 \\
GeoRSCLIP-FT~\cite{georsclip} & 15.26 & 40.46 & 57.79 & 22.14 & 40.53 & 51.78 \\
\midrule
AnySat + Linear Proj. & 12.79 & 36.83 & 52.91 & 16.77 & 34.88 & 48.75 \\
AnySat + Contr. Align. & 15.02 & 41.20 & 58.40 & 20.56 & 39.64 & 53.20 \\
\textbf{\method{}} & \textbf{16.49} & \textbf{43.73} & \textbf{60.35} & \textbf{23.41} & \textbf{42.18} & \textbf{55.67} \\
\bottomrule
\end{tabular}}
\end{table*}

\subsection{RSICD Cross-Dataset Adaptation}

Table~\ref{tab:rsicd} evaluates cross-dataset adaptation on RSICD.
Starting from the BigEarthNet.txt trained model, we fine-tune only the predictive
alignment network on the RSICD training split while keeping the AnySat and
RemoteCLIP encoders frozen. This setting evaluates parameter-efficient adaptation
from Sentinel-based BigEarthNet.txt to RGB aerial image--caption retrieval.
Published RSICD retrieval methods are included for context.

Under the BigEarthNet.txt-to-RSICD adaptation setting, \method{} obtains the strongest results among our evaluated variants. The gains over AnySat + Linear Projection show that direct projection of pooled visual tokens is insufficient for robust language alignment. The improvement over AnySat + Contrastive Alignment suggests that predictive semantic alignment supports effective cross-dataset adaptation across retrieval ranks. These results support the use of AnySat as a unified EO visual encoder for cross-modal and cross-sensor retrieval, while also highlighting that RSICD remains a domain-shifted evaluation setting. The AnySat-based baselines in the bottom block follow the same RSICD
adaptation protocol, with only their trainable alignment components updated.

\subsection{RSVQA Representation Probe}

Table~\ref{tab:rsvqa_comprehensive} reports RSVQA-LR and RSVQA-HR results. We use RSVQA as a representation probe rather than as a generative VQA benchmark. Therefore, the table is organized by evaluation setting: specialist supervised models, generalist or instruction-tuned models, and our representation-probing variants. In our probe setting, only the lightweight answer head is trained on the RSVQA training split, while the image and text encoders remain fixed.

\begin{table*}[ht]
\centering
\caption{RSVQA results under different evaluation settings. Scores are category-specific and average accuracies ($\%$). The bottom section reports our representation probes with a lightweight answer head trained. Rur/Urb, Pres, and Comp denote Rural/Urban, Presence, and Comparison, respectively.}
\label{tab:rsvqa_comprehensive}
\resizebox{\textwidth}{!}{
\begin{tabular}{lccccccc}
\toprule
\multirow{3}{*}{Method} & \multicolumn{4}{c}{RSVQA-LR} & \multicolumn{3}{c}{RSVQA-HR (Test Set 2)} \\
\cmidrule(lr){2-5} \cmidrule(lr){6-8}
 & Rur/Urb & Pres & Comp & Avg. Acc & Pres & Comp & Avg. Acc \\
\midrule
\multicolumn{8}{l}{\textit{Specialist Models (Supervised / Fine-tuned)}} \\
RSVQA~\cite{rsvqa} & 90.00 & 87.46 & 81.50 & 86.32 & 86.26 & 85.94 & 86.10 \\
RSGPT~\cite{rsgpt} & 94.00 & 91.17 & 91.70 & 92.29 & 89.87 & 89.68 & 89.78 \\
\midrule
\multicolumn{8}{l}{\textit{Generalist / Instruction-Tuned Models}} \\
Qwen-VL-Chat~\cite{qwen_vl} & 62.00 & 66.54 & 47.65 & 58.73 & 61.75 & 65.98 & 63.87 \\
GeoChat~\cite{geochat} & 91.09 & 91.09 & 90.33 & 90.70 & 58.45 & 83.19 & 72.30 \\
\midrule
\multicolumn{8}{l}{\textit{Ours (Representation Probes)}} \\
RemoteCLIP + Answer Head & 87.20 & 87.65 & 85.74 & 86.86 & 85.45 & 81.80 & 83.63 \\
AnySat + Answer Head & 86.30 & 89.15 & 88.76 & 88.07 & 84.92 & 78.54 & 81.73 \\
AnySat + Contrastive Alignment + Answer Head & 87.15 & 88.42 & 84.14 & 86.57 & 87.60 & 83.12 & 85.36 \\
\textbf{\method{} + Answer Head} & \textbf{90.38} & \textbf{93.66} & \textbf{91.49} & \textbf{91.84} & \textbf{89.24} & \textbf{89.06} & \textbf{89.15} \\
\bottomrule
\end{tabular}
}
\end{table*}

Among our representation-probing variants, \method{} obtains the highest average accuracy on both RSVQA-LR and RSVQA-HR. The improvement over AnySat + Answer Head shows that language alignment benefits question-conditioned prediction, while the gain over contrastive alignment indicates that the semantic prediction objective improves the usefulness of the learned image-side representation. Since our setup trains an answer head, these results should be interpreted as evidence of representation quality rather than zero-shot or generative VQA ability.

\subsection{Ablation Analysis}

Table~\ref{tab:ablation-summary} summarizes the main ablation results on BigEarthNet.txt. We report mean retrieval performance averaged over text-to-image and image-to-text retrieval. The ablations examine masking policy, loss design, and predictive network design in a single compact table.

\begin{table*}[t]
\centering
\caption{Ablation analysis on BigEarthNet.txt. Scores are mean Recall@K averaged over text-to-image and image-to-text retrieval. Bold indicates the best result.}
\label{tab:ablation-summary}
\resizebox{0.85\linewidth}{!}{
\begin{tabular}{llccc}
\toprule
Factor & Variant & R@1 & R@5 & R@10 \\
\midrule
\multirow{5}{*}{Masking policy}
& No masking & 18.75 & 44.00 & 58.60 \\
& Random, ratio $0.25$ & 19.60 & 45.20 & 60.10 \\
& Random, ratio $0.50$ & \textbf{21.31} & \textbf{48.56} & \textbf{62.87} \\
& Random, ratio $0.75$ & 18.90 & 44.10 & 58.75 \\
& Block, ratio $0.50$ & 20.60 & 47.40 & 61.80 \\
\midrule
\multirow{3}{*}{Loss design}
& Contrastive only & 17.75 & 42.60 & 57.20 \\
& Prediction only & 18.80 & 44.70 & 58.30 \\
& Prediction + Contrastive & \textbf{21.31} & \textbf{48.56} & \textbf{62.87} \\
\midrule
\multirow{4}{*}{Network design}
& Linear projection & 11.20 & 31.15 & 44.10 \\
& Two-layer MLP & 15.80 & 38.70 & 53.00 \\
& Multi-scale predictor + mean pooling & 19.60 & 45.50 & 59.70 \\
& Multi-scale predictor + learned query pooling & \textbf{21.31} & \textbf{48.56} & \textbf{62.87} \\
\bottomrule
\end{tabular}
}
\end{table*}

The ablation results show that the complete design performs best. A moderate random mask ratio of $0.50$ outperforms no masking and heavy masking, indicating partially visible context is useful for predictive semantic alignment. The loss ablation shows that semantic prediction and contrastive retrieval are complementary: prediction-only improves over contrastive-only, but the combined objective performs best. Finally, the network-design ablation shows the benefit of multi-scale prediction and learned query pooling, with the full predictor outperforming linear, MLP, and mean-pooling alternatives.

% ---------------------------------------------------------------
\section{Discussion}

The results suggest that \method{} provides a practical way to align pretrained EO representations with remote-sensing language embeddings without training a full vision-language model from scratch. By keeping the AnySat visual encoder and RemoteCLIP text encoder fixed, only the lightweight predictive alignment network is optimized, which is useful in remote sensing where large-scale multisensor pretraining is costly and performance can vary across modality, geography, resolution, and task~\cite{geobench,pangaea}.

\method{} differs from CLIP-style RS vision-language models by predicting a frozen text-space embedding from partially visible EO tokens rather than relying only on global image--text contrastive alignment~\cite{remoteclip,georsclip}. The ablations indicate that masking, semantic prediction, and contrastive retrieval provide complementary supervision. We interpret the gain from moderate masking as evidence that prediction benefits from distributed contextual evidence rather than a few highly discriminative local tokens. The mask-aware hierarchy further exposes the predictor to complementary fine, regional, and global context while preventing masked features from leaking through coarse representations. RSVQA should therefore be interpreted as a representation probe for question-conditioned prediction, not as evidence of generative VQA or reasoning ability.

The main limitations are domain shift, dependence on the frozen target space, and supervision scope. Because the RemoteCLIP text space is fixed, \method{} performs cross-modal representation mapping rather than compositional language reasoning, and its semantic expressiveness is bounded by the target embedding space. BigEarthNet.txt captions may also not fully capture the visual and linguistic diversity of datasets such as RSICD. Future work can explore longer text targets, phrase-level supervision, referring expressions, and spatial grounding~\cite{lrsclip}.

% ---------------------------------------------------------------
\section{Conclusion}

We presented \method, a JEPA-inspired predictive vision-language alignment
framework for remote-sensing foundation models. By keeping the EO visual and
text encoders fixed and training only a lightweight predictive aligner, the method
maps mask-aware fine, regional, and global visual context into a remote-sensing
text embedding space. This combines hierarchical masked semantic prediction
with bidirectional contrastive retrieval alignment.

Experiments on BigEarthNet.txt, RSICD, and RSVQA demonstrate the potential
of \method{} for natural-language Sentinel retrieval, cross-dataset image--text
adaptation, and question-conditioned representation probing. The retrieval
results consistently improve over the corresponding AnySat-based alignment
baselines, while the ablations show complementary benefits from moderate
visual masking, semantic prediction, contrastive supervision, and learned query
pooling. These findings indicate that structured prediction from partially visible
EO context can provide effective language alignment without updating the
underlying foundation encoders.

Overall, \method{} provides a parameter-efficient route for making EO
foundation model representations more useful for natural-language geospatial
search and analysis. Future work can extend the framework toward richer
language supervision, spatial grounding, and broader sensor and geographic
settings.

% \section*{Acknowledgements}
% The authors would like to thank ...
% \clearpage
% ---- Bibliography ----
\bibliographystyle{splncs04}
\bibliography{main}

\end{document}